# Sequential Planning in Large Partially Observable Environments guided by LLMs

Swarna Kamal Paul
Tata Consultancy Services

*Abstract*— Sequential planning in large state space and action space quickly becomes intractable due to combinatorial explosion of the search space. Heuristic methods, like monte-carlo tree search, though effective for large state space, but struggle if action space is large. Pure reinforcement learning methods, relying only on reward signals, needs prohibitively large interactions with the environment to device a viable plan. If the state space, observations and actions can be represented in natural language then Large Language models (LLM) can be used to generate action plans. Recently several such goal-directed agents like Reflexion, CLIN, SayCan were able to surpass the performance of other state-of-the-art methods with minimum or no task specific training. But they still struggle with exploration and get stuck in local optima. Their planning capabilities are limited by the limited reasoning capability of the foundational LLMs on text data. We propose a hybrid agent "neoplanner", that synergizes both state space search with queries to foundational LLM to get the best action plan. The reward signals are quantitatively used to drive the search. A balance of exploration and exploitation is maintained by maximizing upper confidence bounds of values of states. In places where random exploration is needed, the LLM is queried to generate an action plan. Learnings from each trial are stored as entity relationships in text format. Those are used in future queries to the LLM for continual improvement. Experiments in the Scienceworld environment reveals a 124% improvement from the current best method in terms of average reward gained across multiple tasks.

*Keywords—sequential planning, POMDP, generative AI agent, LLM, state space search*

## I. Introduction

Sequential planning in an environment is to find a sequence of actions or a policy that can meet an objective. As the state space size of the environment increases it becomes increasingly difficult to find a workable plan to a point such that the problem seems intractable with traditional graph or tree based search algorithms. On top of that due to partial observability, the complete state space of the environment may not be determined completely. In such scenario model free reinforcement learning (RL) can help tackle the problem to a certain extent. Model free RL is like a trial-and-error based learning where an optimal policy is learnt guided by rewards provided by the environment. Monte-carlo tree search [1] is such an effective method that has provably tackled very large state spaces. However, its search space can also explode if the action space is large. It will require too many trials to converge to a near optimal policy. This is true for all other pure RL based approaches, that solely relies on reinforcements to determine a policy. In a delayed reward environment, the challenge becomes prominently significant. Due to absence of frequent reward signals the search becomes mostly random for a significant amount of time and that adds to the combinatorial explosion. Purely RL based approaches also face challenges in adapting to sudden changes in the environment.

Humans can adeptly handle intricate planning tasks in real-world scenarios with large state space, large action space and even in partial observability. Suppose a person wants to drink water while she visited her friend's place. She didn't know the water source, but predicts that it might be in the kitchen. Once she finds it, she knows thar she came closer to her goal. After that, she needs to find a way to dispense water into some container, such as a cup and drink it. There can be lot of actions she might take in an unknown home layout, but she would usually achieve her goal in handful number of trials or steps. Human does that as they have extensive knowledge on how common things work, how different objects are related and what actions causes what changes.

Large language models or LLMs pretrained on large amount of real-world text data captures meaningful correlations of words. With sufficiently expressive representation of the problem environment in natural language, the LLM can generate fairly accurate shallow plans. With proper prompting, LLMs can predict sequence of actions in text format that may meet the objective in the environment. However, LLMs are not reasoning machines. They can generate shallow plans reasonably well but may fail to do deep reasoning and generate a lengthy policy. They also get stuck in local minima due to their greedy approach and fails to explore significantly. For example, if initially the human starts from outside, she doesn't observe there is a kitchen in the home. To find a drinkable water source she needs to explore and find the kitchen and then the water source. A LLM based agent might instead try to locate irrelevant water sources in its current observation range, like wet clothes tied outside.

In short, pure RL based approaches face the problem of combinatorial explosion. Pure LLM based approaches face the problem of shallow reasoning and lack of exploration. We propose to combine the best of two approaches in a novel way to alleviate both the problems. We propose to build a state space model of the environment by trying out different actions, recording observations and rewards. The model would provide an anytime best policy, based on what is known at any moment. Wherever random exploration of actions is needed, the LLM is used to predict the best sequence of actions. We incrementally build a memory of learnings about the environment based on all previous trials. The accumulated learnings help the LLM to more accurately predict the sequence of actions in future. Building a state space model guarantees the exploration of the environment and keeps a balance of exploration and exploitation. The use of LLM helps to dampen the combinatorial explosion of action space and state space where random exploration is needed.

We tested our method in Scienceworld [2] environment. It is collection of scientific tasks, all represented in natural language. The environment has huge number of possible actions and objects, making it a huge state space. The environment is partially observable, such that the state information is not explicitly available. The latent state needs

Author Email - swarna.kpaul@gmail.com

to be derived from the observations. The full environment is not visible initially. Experimental results reveal our developed agent completed or nearly completed all tasks and surpassed all prior state of the art methods by a wide margin.

## II. RELATED WORKS

There is a rich literature on sequential planning in large environments. DRRN [3], KGA2C [4] and CALM [5] are deep reinforcement learning based approaches to solve large text-based environments. They usually require very large number of interactions with the environment before coming up with a workable plan. Delayed reward in the environment amplifies the problem further. There are another class of planning algorithms for large POMDPs, called as direct search algorithms [11,12,13]. They don't rely much on the state space information but do a search in solution space. But they also can have very high complexity if the action space or solution size is large. LLM based agents like SayCan [6], ReAct [7], Reflexion [8], CLIN [9] exploits the large pretraining of an LLM with real-world semantic knowledge and uses it to suggest next best actions without further significant training. These approaches use intelligent step-by-step prompting tricks and maintains problem specific memories to be used by the LLM to optimize action selection. However, these methods are susceptible to premature convergence to suboptimal pathways and struggle with exploration. These methods are limited by the depth of reasoning needed to choose an action as LLMs have limited capability of reasoning on text at one go.

## III. APPROACH

The problem environment – ξ, is modelled as deterministic Partially Observable Markov Decision Process (POMDP) [10]. Many real-world problems can be formulated in this model. The environment can be modeled as the following tuple: $(S, A, O, s_0, T, \mathbb{R}, f(s,a), o(s,a), r(s,a), \tau)$.

$S \rightarrow$ a finite set of states, $A \rightarrow$ a finite set of actions, $O \rightarrow$ a finite set of observations, $s_0 \rightarrow$ initial state, $T \rightarrow$ goal state, $\mathbb{R} \rightarrow$ set of rewards, $f(s,a) \in S, for\ s \in S, a \in A \rightarrow$ deterministic state transition function after applying an action, $o(s,a) \in O, for\ s \in S, a \in A \rightarrow$ deterministic observation function after applying an action, $r(s,a) \in \mathbb{R}, for\ s \in S, a \in A \rightarrow$ reward function for getting reward after applying an action, $\tau \rightarrow$ policy or sequence of actions that might meet the objective.

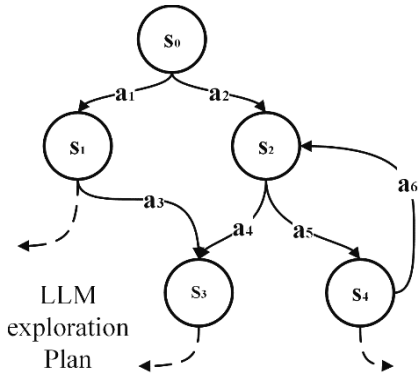

**Fig.1** Illustration of learnt state space model and planning by LLM where random exploration is needed

Fig.1 illustrates the state space model building and searching for the optimal policy. The agent starts from some initial state $s_0$. As states are not directly accessible from the environment, the observations can be encoded in certain way to derive the latent state. Each state is assigned a value augmented with an exploration term, similar to one used in UCB1 [14]. The agent chooses the path to maximize the upper confidence bound of the value of states. The action plan is generated based on the selected pathway. If a leaf node is reached or a random exploration needed in a node that is yet to be fully expanded, a foundational LLM is prompted to generate rest of the action sequence.

Once an action plan is generated, it is executed in the environment. The perceived observations, states and rewards are used to update the state space graph and the value of the nodes. The observations and rewards are also used to update a consolidated learnt memory in text format that can be further used by the LLM for future action planning. Instead of random selection of actions for exploration, the LLM can provide a better educated guess on the actions required to reach the goal state, thereby damping the search space.

## IV. STATE SPACE GRAPH BASED PLANNING

The agent takes actions in the environment and based on the perceptions received it builds a state space graph of the environment. The rewards are used to update the values of the states. The values of the states are updated using Temporal Difference (TD) learning. The state space graph is eventually used to determine the optimal policy. The complete planning process can be divided into following major steps. Namely, plan selection, plan exploration, simulation, state space learning.

### A. Plan selection

Each state in the state space graph is assigned a value $V(s)$ based on the future rewards it receives from the current state. Thereafter an augmented value is calculated for each state by the by adding an exploration term to $V(s)$, where $C$ is a constant, $N$ is total number of simulations of all the parent states of $s$ and $n_s$ is total number of simulations of state $s$.

$$V^{\oplus}(s) = V(s) + C\sqrt{\frac{\ln N}{n_s}} \qquad (1)$$

From current state the action is selected that produces a valid child state with maximum $V^{\oplus}$ in the state space graph. This selection continues until a leaf node is reached or maximum $V^{\oplus}$ of the child state is less that the default exploration $V^{\oplus}$ or a closed loop is detected. The selected path is converted to sequence of actions that serves as the first part of action plan. Along with the action plan, a list of actions to be avoided from the current state (if any) is also returned. If the final state in the returned action plan has other child nodes, then the actions corresponding to those edges are returned as actions to be avoided. Those actions can be avoided because they might either be invalid actions or lead to states that are already explored and have not enough values to be explored further. The list of actions to be avoided and the current state are properly annotated in text format and returned as additional instructions ($\Phi$) for the LLM to guide the search through action space.

### B. Plan exploration

The formula of $V^{\oplus}$ as stated in equation 1 captures both the metrices for exploitation and exploration. The value of a state depicts how much the state can be exploited to reach towards the goal state. The second term in $V^\oplus$ measures how less a state and its children states has been explored. $V^\oplus$ maintains a balance between exploration and exploitation in the state space graph. But this will work only for the visited or known states that are available in the state space graph. As the agent starts to interact with the environment, most of the states remain unexplored. There needs to be a way to select actions and explore the unexplored states. A simple random policy may quickly become intractable if action space is large. To explore the unexplored states the LLM is prompted to generate an action plan from the current state. For each state a default explore $V^\oplus$ is calculated by the following formula, where $V_{default}$ is default value of an unexplored state, $K_s$ is a default exploration factor for state $s$.

$$V^\oplus_{explore}(s) = V_{default} + K_s C \sqrt{\ln n_s} \qquad (2)$$

For each state $s$, a child state $s'$ is selected that has the maximum value of the following metric.

$$V^\oplus_K(s') = V(s') + K_s C \sqrt{\frac{\ln N}{n_{s'}}} \qquad (3)$$

If $V^\oplus_{explore}(s) > V^\oplus_K(s')$ then the subsequent action plan is generated using the LLM from the current state $s$.
The factor $K_s$ estimates amount of exploration needed from the current state based on how many un-tried actions are left. $K_s$ is calculated as following, where $|a_s|$ is total number of possible actions from state $s$, $|\overline{a_s}|$ is total number of actions already taken in state $s$ and $n$ is a non-linearity factor greater than 1.

$$K_s = \left(\frac{|a_s| - |\overline{a_s}|}{|a_s|}\right)^n \qquad (4)$$

$K_s$ ensures that exploration gets reduced as greater number of actions are tried from a state and eventually drops to 0 if all actions are tried out.

*C. Simulation*

Once an action plan is generated it is executed in the environment in sequence. For each action the corresponding observation is recorded as action-observation sequence ($x_{AO}$) and state information is recorded as action-state sequence ($x_{AS}$). The corresponding reward for each action and total number of available actions from a state is also recorded in $x_{AS}$. Eventually this sequence is used to update the state space graph. $x_{AO}$ is used to update the learnings in text format. If an action is invalid then it is marked as invalid in the $x_{AS}$.

*D. State space learning*

After simulation the state space graph is updated using $x_{AS}$. For each action-state sequence the start state and end state are searched in the state space graph. If an edge between the states does not exists, the edge is added. If the state nodes do not exist the nodes are created. If the edge already exists then corresponding state visit count is incremented. Each node in the state space graph stores the following items.
$s \rightarrow$ encoded description of the state, $V(s) \rightarrow$ value of the state, $n_s \rightarrow$ number of visits in the state, $|a_s| \rightarrow$ total possible actions from the state.

Each edge contains the following items.
$a \rightarrow$ action name, $r(s, a) \rightarrow$ reward obtained by taking the action, $s_s \rightarrow$ start node, $s_e \rightarrow$ end node.
If an action is invalid then the sink of the edge goes to a fixed invalid node ($s_\otimes$) in the graph. The graph starts with a fixed root node and the initial state is connected to the root node. Once the required nodes and edges are created from $x_{AS}$ the values of all nodes are updated using TD learning. The following TD(0) update is applied for each state in the state space graph to update the value, where $\alpha$ is a constant step size parameter and $\gamma$ is constant decay factor.

$$V(s) = V(s) + \alpha[r(s, a) + \gamma V(s') - V(s)] \qquad (5)$$

This update is carried out for $i$ iterations, in the expectation that the values of the states would converge. The update is interrupted in between if the average change in values of all states is below a threshold in an iteration. After completing the value update process, the $V^\oplus$ gets updated for all states as per equation 1. The factor $K_s$ also gets updated for each state as per equation 4.

## V. USING LLMs AS PROBABILISTIC ORACLE

Large language models or LLMs are deep neural networks, mostly based on transformer architecture, are trained on large corpus of text data. They can be used for text completion for a given input text snippet. Recently, LLMs like GPT4, was able to generate human like responses for many input texts on wide variety of subjects. Due to its rigorous and extensive real-world training data, it is able to capture lot of real-world semantics in its model in the form of correlation among words. Even though it just does next word prediction, but due to attention mechanism the predicted next words are generally semantically aligned with the whole meaning of the prior text.

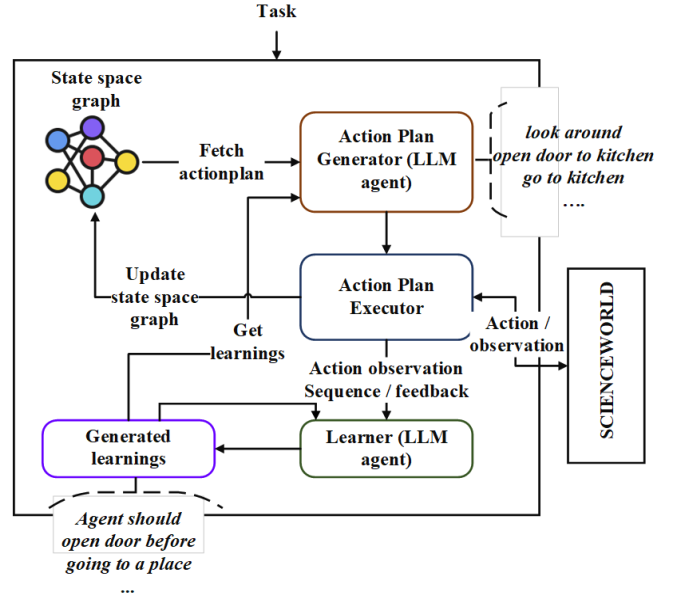

Fig. 2 Architecture of the planner.

If everything about the problem environment can be represented in text form (states, actions, observations, feedback), LLMs can be used to generate a sequence of actions. Fig.2 represents the architecture of the agent. As mentioned in Section IV, the LLM is used to generate an

**Algorithm**

**procedure** solve($\xi$)
  initialize $SG$ as state space graph
  **while** True
    **if** goal reached
      break
    $x_{AO} = []$
    $x_{AS} = []$
    **for** i in 1 to j:
      $\Phi, \tau$ = selectplan($SG, s_\xi$)
      $\tau'$ = generateactionplan($\Phi, x_{AO}, \xi$)
      $\tau = \tau + \tau'$
      $\overline{x_{AO}}, x_{AS}$ = executeactionplan($\tau, \xi$)
      updatestatespacegraph($SG, x_{AS}$)
      $x_{AO} = x_{AO} + \overline{x_{AO}}$
    $f$ = getfeedback($x_{AS}$)
    learner($x_{AO}, f, \xi$)
    reset($\xi$)

**procedure** selectplan($SG, s_\xi$)
  set parent state $s_p = s_\xi$
  **while** True
    $S_c$ = getchildstates($SG, s_p$)
    **if** $S_c = \emptyset$
      break
    $s_c, a_c = \underset{s,a}{\operatorname{argmax}} V^\oplus(S_c)$
    $K_{s_p} = (\frac{|a_{s_p}| - |\overline{a_{s_p}}|}{|a_{s_p}|})^n$
    $V_K^\oplus(s_c) = V(s_c) + K_{s_c} C \sqrt{\frac{\ln N}{n_{s_c}}}$
    $V_{explore}^\oplus(s_p) = V_{default} + K_{s_p} C \sqrt{\ln n_{s_p}}$
    **if** $V_{explore}^\oplus(s_p) > V_K^\oplus(s_c)$ or $\forall S_c = s_\otimes$
      add actions leading to $S_c$, as list of actions to be avoided in $\Phi$
      break
    **if** state space loop detected by adding $a_c$ in $\tau$
      break
    add $a_c$ in $\tau$; set $s_p = s_c$
  add $s_p$ in $\Phi$
  **return** $\Phi, \tau$

**procedure** generateactionplan($\Phi, x_{AO}, \xi$)
  extract $t_\xi, d_\xi, l_\xi$ from $\xi$ and add in $\Phi$
  $P_{explore} = (\frac{\sigma}{\ln N_{exp}})$
  update $t_\xi$ in $\Phi$ by $t_{eo}$ with probability $P_{explore}$
  add $x_{AO}$ in $\Phi$
  $\tau$ = LLMpredict($\Phi$)
  **return** $\tau$

**procedure** updatestatespacegraph($SG, x_{AS}$)
  **for** $a, s, r(s, a)$ in $x_{AS}$
    add $s$ as node in $SG$ if not present
    add or update $a, r(s, a)$ as edge in $SG$
  **for** 1 to $i$
    **for** $s_p$ in all nodes in $SG$
      **for** $s_c$ in all valid child nodes of $s_p$
        $a_c \leftarrow$ action from $s_p$ to $s_c$
        $V(s_p) = V(s_p) + \alpha[r(s_p, a_c) + \gamma V(s_c) - V(s_p)]$
  **for** $s$ in all nodes in $SG$
    $V^\oplus(s) = V(s) + C\sqrt{\frac{\ln N}{n_s}}$
    $K_s = (\frac{|a_s| - |\overline{a_s}|}{|a_s|})^n$

**procedure** getfeedback($x_{AS}$)
  extract reward set $R$ from $x_{AS}$
  $\mathbb{R} = \sum R$
  $f$ = annotate $\mathbb{R}$ to text feedback
  **return** $f$

**procedure** learner($x_{AO}, f, \xi$)
  initialize $\Phi$ as learner prompt
  extract $t_\xi, l_\xi$ from $\xi$ and add in $\Phi$
  add $x_{AO}$ and $f$ in $\Phi$
  $l_\xi$ = LLMpredict($\Phi$)
  set $l_\xi$ as updated learnings in $\xi$

exploration plan where a random exploration is needed in the state space graph, due to unavailability of no further state space information. The action plan is generated by the action plan generator. Before generating an action plan, it queries the state space graph to select the best plan as per the method mentioned in section IV-A. The rest of the action plan is generated from the final state of the selected action plan. The generated action plan is appended with the selected action plan from the state space graph. The final action plan is executed in the environment by the action plan executor. It carries out the simulation step mentioned in Section IV-C. The action plan generator and action plan executer are run for several iterations in sequence, after which the environment is reset. Each reset marks end of an episode and begin of a new one. After each episode, the recorded action-observation sequence along with previous learnings (if any) are fed into a Learner agent. It generates a set of learnings about the environment in free text format.

*A. Action Plan Generator*

The action plan generator generates a sequence of actions that might meet the objective ($t_\xi$) in the environment from the current given state. It takes the objective, a prior description of the environment ($d_\xi$), current state description ($s_\xi$), previous learnings about the environment ($l_\xi$), action-observation trace ($x_{AO}$), list of actions to be avoided from the current state, as parameters and generates a text prompt. The text prompt is sent to LLM to get an action plan ($\tau'$). If the output is not generated in correct format it is resend to the LLM again. The objective and prior description of the environment is set during initialization of the task. Intermittently the task objective is replaced with the following text (*exploration objective - $t_{eo}$*) to promote exploration and enrich the learnings.

*"Create a long sequence of actions to explore and know more about the environment".*

In each run of action plan generator, the task objective is replaced with exploration objective by the following probability, where $\sigma$ is a constant and $0 < \sigma < 0.5$, $N_{exp}$ is total number of times exploration objective has been run. We set $\sigma = 0.3$.

$$P_{explore} = (\frac{\sigma}{\ln N_{exp}}) \qquad (6)$$

The current state description is obtained from the state-space graph. The current state is the final state of the selected action plan from the state space graph. The list of actions to be avoided is also obtained from the state space graph.

## B. Learner

The learner is an LLM agent that generates learnings about the environment. It takes the $x_{AO}$ trace of an episode, the previous learnings, feedback after running the last episode, as parameters and creates a prompt for the LLM. The LLM generates updated learnings as list of text. The learnings are generated in the format *"X Y Z"*, where X and Z are entities, subject, object, events from $x_{AO}$ trace and Y is relation between X and Z. This captures all the important relationships of entities in the environment that can be used to meet the objective.

## VI. EXPERIMENTAL RESULTS

We opted for ScienceWorld [2], an interactive text-based environment that demands intricate interactive reasoning processes for resolving a multitude of science-theory-based tasks across various classes, such as thermodynamics, genetics, friction, and more. This virtual space encompasses ten sub-locations: foundry, greenhouse, outside area, art studio, workshop, kitchen, living room, bedroom, bathroom, and a hallway connecting internal spaces. The complexity of the environment, characterized by multiple objects, their respective states, and action templates, results in an intractable search space for any agent. Both the action space and state space is huge. The complete state space of the environment is not visible at the beginning. There are no state information available from the environment. It needs to be derived from the observations.

We developed neoplanner in Python3. The code is available at https://github.com/swarna-kpaul/neoplanner. We tested the agent in 7 different environments. For each environment the agent runs for several episodes. At the start of each episode the environment is reset and agent is initialized at the same start location. During an episode the agent tries multiple actions and cumulative reward is calculated. The total reward is transformed into natural language feedback in same way as done in [9]. The reward is log transformed before adding into $x_{AS}$, that is eventually used to update state space graph. We used the GPT4-turbo model as LLM. Table I shows the total reward gained by different methods across multiple environments. Our proposed method surpasses all state of the delayed reward setting, where a reward is obtained only after performing several actions towards the objective. For example, the *"boil"* task is a delayed reward task. It receives very infrequent rewards and rewards are obtained only after performing several actions towards boiling water. The RL methods performed poorly for this task as they barely found any signals to traverse through large state space. The generative language based agents performed better in these tasks. As they capture lot of real-world semantics, with proper description of the environment, and continual learnings, they are able to select better action plans. However, these methods mostly ignore the numerical reward signals and rely on textual feedbacks. This causes rapid convergence to suboptimal plans and the agent struggle with exploration [9]. Our proposed agent used both generative language based action planning and policy generation through state space graph search. This helps to keep a balance between exploration and exploitation without drowning into combinatorial explosion during exploration.

Most of the generative language based agents generate action plans by finding the next best action. Thus, they have to query the LLM every time an action needs to be taken. On a contrary, neoplanner queries the LLM to find a sequence of actions at one go, such that the number of calls to the LLM can be reduced. This potentially reduced the cost of generating the action plan.

The agent started with a blank memory with no learnings. While solving a task it generated several task related learnings about the environment. For example, in the "use-thermometer" task it generated following learnings among many other.

*"thermometer in kitchen can be moved to inventory"*, *"agent can move between kitchen, living room, hallway, art studio, greenhouse, bedroom, workshop"*

The same set of learnings are used to bootstrap for the next task. For solving the next task *"measure-melting-point-known-substance"* some of the learnings are useful, but many are not useful. While solving this task it updated its learnings related to the current task. It generated some learnings like *"picking up chocolate from fridge and focusing on it is part* art methods by wide margin. *#interactions* for neoplanner depicts the total number of interactions done with the environment. The metrices for all state of the art methods have been taken from [9].

TABLE I. Comparing neoplanner against SOTA

| Task | RL Methods | | | Generative Language Agents | | | | NeoPlanner |
|---|---|---|---|---|---|---|---|---|
| | **DRRN** | **KGA2C** | **CALM** | **SayCan** | **ReAct** | **Reflexion** | **CLIN** | **(#interactions)** |
| use-thermometer | 6.6 | 6.0 | 1.0 | 26.4 | 7.2 | 5.9 | 25.2 | **85** (390) |
| measure-melting-point-known-substance | 5.5 | 11.0 | 1.0 | 8.0 | 6.1 | 28.6 | 58.2 | **100** (192) |
| find-plant | 15.0 | 18.0 | 10.0 | 22.9 | 26.7 | 64.9 | 100 | **100** (227) |
| chemistry-mix | 15.8 | 17.0 | 3.0 | 47.8 | 51.0 | 70.4 | 51.7 | **100** (693) |
| biology-identify-life-stages-1 | 8.0 | 10.0 | 0.0 | 16.0 | 8.0 | 8.0 | 32.0 | **92** (515) |
| boil | 3.5 | 0.0 | 0.0 | 33.1 | 3.5 | 4.2 | 16.3 | **100** (81) |
| freeze | 0.0 | 4.0 | 0.0 | 3.9 | 7.8 | 7.8 | 10.0 | **82** (154) |
| **Average Reward** | 7.8 | 9.4 | 2.1 | 22.6 | 15.8 | 27.1 | 41.9 | **94** |

### A. Discussion

Pure RL methods can be effective sometimes, but as they focus mostly on the reward signals, the number of training trials required can be enormous. The problem amplifies in

*of the task"*. Thus, the agent is capable of adapting its learnings.

### B. Scope of improvements

Though the agent completed the objective for many of the tasks and received maximum reward (100), yet for few tasks, it didn't follow the ideal shortest action plan. Several irrelevant actions (that didn't cause changes in reward) are chosen as part of the action plan. For example, in the *"boil"* task, the actions "*pick up sodium chloride*" and "*mix sodium chloride*" are part of the plan but they are irrelevant. This is because it just moved from states to states during exploration, and followed the rewards. The actions that yielded rewards came after the above 2 actions. A state space pruning strategy may be used to handle this scenario. Irrelevant states within the action plan maybe identified in the states space graph and pruned.

The plan generated is not the ideal and shortest one. In the "*boil*" task, the agent turned on the stove containing metal pot with water, waited for several iterations, and directly focused on the steam. The ideal plan would be to measure the temperature of the water every certain interval. If it exceeds $100^oC$ then the task is complete. This problem can be tackled by representing policies as programs in a programming language instead of just sequence of action texts. Representing policies as programs would allow to implement all type of logical and mathematical constructs in the policy. That would allow implementation of reasoning within the policy.

As the agent starts solving a new task with learnings from previous task, the LLM generates action plans tuned for the previous task and it takes a while to update the memory and adjust the generation of action plan for new task. This happens due to presence of irrelevant learnings for the current task (relevant for the previous). For the "*boil*" and "*freeze*" tasks, this problem was prevalent and the action plan was not converging. So, we took the memories from "*chemistry-mix*" and ran few episodes of "*boil*" and "*freeze*" alternately to come up with a common relevant memory. This problem can be alleviated by a better memory management strategy, so that only relevant memories for the current task are fetched.

## VII. CONCLUSION

Combining RL method of searching state space and querying LLM to generate an action plan have proven to be effective in solving large environments with a relatively smaller number of interactions with the environment. Neoplanner constructs the state space graph and updates the value of the states as it interacts with the environment. The same state space graph is exploited to find action plans progressively. It also generated learnings about the environment in text format as it continued interaction. The learnings helped in continuous improvement of the action plan. The agent demonstrated adaptability of the learnings as the task changed. The learnings also generalized as the number of trials progressed. Neoplanner was able to completely solve several tasks and nearly solve rest of them. With better memory management of the learnings, the agent could converge towards the objective even faster. Also representing the plan as a program can help to reduce the plan size and make it more general to apply on other similar tasks.

## APPENDIX

Following are the prompts used for action plan generator and the learner.

---

**Action plan generator prompt**

System: You are an AI action planner for an autonomous agent. You are situated in a task environment, as provide by the user, prior axioms are the fixed rules and constraints of the environment, the belief axioms are your beliefs about the environment. You need to generate an action plan that will contain a sequence of actions to meet the objective of the environment. Do not generate any additional explanations. The output should be a list of actions.

Here are some example outputs.
  {actionplanexamples}
Here are some historical traces of action and observation:
{envtrace}

User: Generate the action plan for the following environment. If there are ADDITIONAL INSTRUCTIONS, then give MOST IMPORTANCE on the ADDITIONAL INSTRUCTIONS to generate the action plan.
Environment:
    {environment}
ADDITIONAL INSTRUCTIONS:
    {instructions}

---

## Learner prompt

System: You are an expert assistant. You are given ACTION OBSERVATION TRACE, a sequence of actions that an agent made in an environment to accomplish a task and the perceptions it got.

You need to derive a comprehensive LEARNINGS as BELIEFAXIOMS. Capture all the details in the ACTION OBSERVATION TRACE.

You can use the beliefaxioms from a list of related similar problem environments to derive the new one.

Generate beliefaxioms, that will help the agent to successfully accomplish the SAME objective AGAIN, in the SAME environment.

Each line can ONLY be of the following forms:
X Y Z

where X and Z are entities, subject, object, events from action perception trace and Y are relation between X and Z. DO NOT add "_" in X, Y or Z. Rigorously capture everything in the action observation trace as memory.

Update on top of the current estimated belief axioms of the current environment based on the action observation trace. Do not remove the existing beliefs.
Modify or remove the existing beliefs only if it contradicts with ACTION OBSERVATION TRACE. You can add your new beliefs to the belief axioms.

The output should always be STRICTLY generated in the following list structure. Each element of list will be a text enclosed in DOUBLE QUOTES. add proper escape characters in the text if required. DO NOT enclose the list in ``` tags.
[ <list of learnings. do not write redundant or contradicting statements> ]

Here is the environment objective and current belief axioms. You should update and output the belief axioms based on the action observation trace provided by the user.
Environment:
    {environment}

User: Here is the action observation trace. Provide the belied axioms for this. COMBINE MULTIPLE LINES OF BELIEFAXIOMS INTO ONE, WHEREVER POSSIBLE.
Action observation trace:
    {envtrace}

Here is the feedback on the overall progress of the agent
    {feedback}

AI:

Here are some examples of the parameters provided in the prompts.

**actionplanexamples** - Example 1:
    ["look around", "open door to greenhouse"]
Example 2:
    ["go door to hallway", "open door to kitchen"]

**environment** - objective:
Your task is to measure the temperature of unknown substance B, which is located around the living room. First, focus on the thermometer. Next, focus on the unknown substance B. If the unknown substance B temperature is above 100.0 degrees Celsius, place it in the red box. If the unknown substance B temperature is below 100.0 degrees Celsius, place it in the green box. The boxes are located around the bathroom.

prior axioms:
    an agent situated in textual task environment. Generate a sequence of actions to meet the objective.
    you may reset the environment if you feel stuck and need to start over.
    FOCUS is an extremely critical action that can be only used the number of times 'focus' is mentioned in the task description and in the exact same sequence. Using it more than that or inappropriately (such as on a wrong object) will terminate the session and the task will be rendered as incomplete. focus can be used on the object which is available in current state.
    Do not make up new actions or objects. If some events need some time to occur after some action is taken then take the action wait to observe the effect after some time.

    DO NOT TAKE ANY ACTION ON ANY OBJECT that is NOT IN ACCESSIBLE OBJECTS in CURRENT STATE

    Here are the following set of allowed actions. where OBJ should be replaced by any object that you can find in your current state.
    Set of parameter values:
    ['activate OBJ', 'close OBJ', 'connect OBJ to OBJ', 'deactivate OBJ', 'disconnect OBJ', 'dunk OBJ in OBJ', 'eat OBJ', 'flush OBJ', 'focus on OBJ', 'go OBJ', 'inventory', 'look around', 'look at OBJ', 'look in OBJ', 'mix OBJ', 'move OBJ to OBJ', 'open OBJ', 'pick up OBJ', 'pour OBJ in OBJ', 'put down OBJ', 'read OBJ', 'reset task', 'task', 'use OBJ on OBJ', 'wait', 'wait1']

    belief axioms:
hallway contains agent, substance called air, picture; has closable doors to art studio, bedroom, greenhouse, kitchen, living room, workshop; door to living room can be opened by agent,
living room contains chair, couch with white pillow, desk with drawer, painting depicting lush green meadow by Bob; has door to hallway; door can be opened by agent',
kitchen contains chair, counter with bowl containing red apple, banana, orange, potato and drawer, cupboard with closed door, freezer with closed door, fridge with closed door, glass jar with substance called sodium chloride, lighter, oven which is off with closed door, painting, sink which is off, substance called soap, deactivated stopwatch, stove which is off, table with glass cup, thermometer reading 10 degrees Celsius; has door to bathroom which is closed, door to hallway which is open, door to outside which is closed,
agent can move between connected rooms

**instructions**: You are at the state:
Currently you see the following things:
This room is called the hallway. In it, you see:
the agent
a substance called air
a picture
You also see:
A door to the art studio (that is closed)
A door to the bedroom (that is closed)
A door to the greenhouse (that is open)
A door to the kitchen (that is open)
A door to the living room (that is closed)
A door to the workshop (that is closed)

Currently you can access the following objects:
['agent', 'air', 'art studio', 'art studio door', 'bedroom', 'bedroom door', 'door to greenhouse', 'door to kitchen', 'door to living room', 'door to workshop', 'greenhouse', 'hallway', 'kitchen', 'living room', 'picture', 'workshop']

The agent has following things in its inventory.

a banana
an orange

find rest of the action plan. You should STRICTLY AVOID the following IMMEDIATE ACTIONS from the current state.
go greenhouse
go living room
focus on thermometer

**feedback:** The agent performed poorly and made some progress but not enough to solve the task.